\documentclass[10pt, a4paper, conference, compsocconf]{IEEEtran}
\IEEEoverridecommandlockouts
\usepackage{url}
\usepackage{graphicx}
\usepackage{subfigure}
\usepackage{multirow}

\ifCLASSINFOpdf
\else
\fi
\begin{document}
\title{Effective Data Augmentation Approaches to End-to-End Task-Oriented Dialogue}


\author{\IEEEauthorblockN{Jun Quan and Deyi Xiong$^*$ \thanks{$\;\;$ $^*$Corresponding author}}
\IEEEauthorblockA{School of Computer Science and Technology\\
Soochow University\\
Suzhou, China\\
terryqj0107@gmail.com;\quad dyxiong@suda.edu.cn}
% \and
% \IEEEauthorblockN{Authors Name/s per 2nd Affiliation (Author)}
% \IEEEauthorblockA{line 1 (of Affiliation): dept. name of organization\\
% line 2: name of organization, acronyms acceptable\\
% line 3: City, Country\\
% line 4: Email: name@xyz.com}
}


%Telephone: (800) 555--1212, Fax: (888) 555--1212}


\maketitle

\begin{abstract}
The training of task-oriented dialogue systems is often confronted with the lack of annotated data. In contrast to previous work which augments training data through expensive crowd-sourcing efforts, we propose four different automatic approaches to data augmentation at both the word and sentence level for end-to-end task-oriented dialogue  and conduct an empirical study on their impact. Experimental results on the CamRest676 and KVRET datasets demonstrate that each of the four data augmentation approaches is able to obtain a significant improvement over a strong baseline in terms of Success F$_1$ score and that the ensemble of the four approaches achieves the state-of-the-art results in the two datasets. In-depth analyses further confirm that our methods adequately increase the diversity of user utterances, which enables the end-to-end model to learn features robustly.

\end{abstract}


% \fi
\IEEEpeerreviewmaketitle

\section{Introduction}
Task-oriented dialogue systems have evolved from traditional modularized pipeline architectures \cite{rudnicky1999creating,zue2000juplter,zue2000conversational} to recent end-to-end trainable frameworks \cite{eric2017copy,eric2017key,lei2018sequicity}. However, the major data challenge for both formalisms remains the same: the lack of annotated dialogue dataset in specific tasks or domains. Various slots and values in dialogue utterances need to be manually labeled for use in supervised learning. As the process of manual annotation is time-consuming and expensive, publicly available task-oriented dialogue datasets normally contain only a few thousand dialogues. For data-driven dialogue systems, especially neural dialogue systems which are more data-hungry, insufficient training data will substantially limit their power to learn from data, resulting in poor robustness and performance.

In this paper, we are interested in handling such a data scarce problem via automatic and cheap data augmentation methods. We propose four different data augmentation approaches: synonym substitution and stop-word deletion at the word level, translation and paraphrasing at the sentence level. We only apply these approaches to rephrase user utterances while keeping machine utterances intact on the training data. For user utterances, we leave slots and corresponding values unchanged and reword the remaining parts, keeping the meanings of user utterances as much the same as possible. In doing so, we hope to diversify user utterances so that our dialogue system can learn to deal with language variabilities in a robust way.

We use TSCP, an end-to-end dialogue system,  recently proposed by \cite{lei2018sequicity} to validate the efficacy of our methods. We conduct experiments on two public datasets, CamRest676 and KVRET. The combination of the four data augmentation methods can collectively outperform the basic TSCP model by 4.5 points in terms of F$_1$ score, the TSCP model with reinforcement learning (RL) by 2.5 points on the CamRest676 dataset. Higher improvements are achieved on the KVRET dataset, 7.8 points and 4.1 points in terms of F$_1$ over the basic TSCP model and TSCP+RL respectively.

The contributions of the paper are threefold: 
\begin{itemize}
\item First, we present and empirically investigate four different approaches to data augmentation for end-to-end task-oriented dialogue, which, to the best of our knowledge, is the first attempt in automatic data augmentation for task-oriented dialogue. 
\item Second, we achieve the state-of-the-art performance on the two datasets with the proposed methods. 
\item Third, our analyses further display that data augmentation on user utterances is better than augmentation on machine utterances. Details on how the proposed methods improve the performance are also provided. 
\end{itemize}

\section{Background: End-to-End Task-Oriented Dialogue}
Task-oriented dialogue systems that can be trained end-to-end have been studied in recent years as alternatives to traditional pipeline-style dialogue systems. Without loss of generality, we use Sequicity \cite{lei2018sequicity} as our baseline system to evaluate our data augmentation methods. It significantly outperforms state-of-the-art pipeline-based methods and obtains a satisfactory entity match rate on out-of-vocabulary (OOV) cases where pipeline-designed competitors totally fail. Sequicity handles both task completion and response generation in a single seq2seq model which can be further optimized with reinforcement learning. It provides a theoretically and aesthetically appealing framework, as it achieves true end-to-end trainability with one single seq2seq model. The key concept introduced in Sequicity is the belief span (bspan), a text span that tracks the dialogue belief states at each turn.

Based on this concept, Sequicity decomposes the task-oriented dialogue problem into the generation of bspans and machine responses in a seq2seq framework. Specifically it decodes in two stages. In the first stage, it generates a bspan to facilitate knowledge base (KB) retrieval. It then generates a machine utterance in the second stage, conditioned on the knowledge base search result and the bspan from the previous stage. Our work is based on an implementation of the Sequicity as a two-stage copynet (TSCP). In the implementation, CopyNet \cite{gu2016incorporating} is used to instantiate Sequicity to allow key words from previous utterances to recur in bspans and generated machine responses.

\section{Data Augmentation Approaches}
In this section, we elaborate the four data augmentation approaches at both the word and sentence level. 

\subsection{Word-Level Data Augmentation}

We substitute words with their synonyms and delete stop words so as to produce diversity in user utterances at the word level.

In synonym substitution, we first utilize the NLTK toolkit \cite{bird2004nltk} and WordNet \cite{miller1990introduction,d2015robust} to conduct part-of-speech tagging and synonym retrieval respectively. In order to ensure that the meaning of user utterances does not change semantically, we only allow some specific words to be replaced by their synonyms. Proper nouns (e.g., \emph{Africa, America}), qualifiers (e.g., \emph{the, a, some, most, every, no}), personal pronouns (e.g., \emph{hers, herself, him, himself}), and modal verbs (e.g., \emph{can, cannot, could, couldn't}) should not be replaced as the substitution of them can easily result in inconsistent statements or even semantic changes. For notional verbs (e.g., \emph{want, like, tell, find}), adjectives (e.g., \emph{cheap, great, delicious}) and nouns (e.g., \emph{food, restaurant, area, south}), we look up their synonyms from WordNet and select the candidate synonyms whose part-of-speech tags are consistent with the corresponding words. For each user utterance, we randomly sample one word that satisfies our substitution rules and randomly select a synonym candidate to replace it. In this way, multiple user utterances can be randomly generated for each original utterance in the training data. These generated utterances will be added to the training data to increase diversity at the word level.

Similarly, we can obtain varieties by deleting stop words in user utterances without changing their meaning. It is common for users to ignore stop words, such as articles, prepositions, adverbs and conjunctions. In order to improve the robustness of the task-oriented dialogue system, and to enable it to pay more attention to the key semantic information in user utterances, we propose to discard these high-frequency stop words in user utterances.

\subsection{Sentence-Level Data Augmentation}
For data augmentation at the sentence level, we investigate two different approaches: translation and paraphrasing. These two methods will improve the sentence-level variances, not limited to the presence/absence or variety of some specific words.  

We use neural machine translation (NMT) models to translate user utterances into other languages and then use reversed NMT systems to translate the generated translations from other languages back to the original language.  In this paper, we use Google online translation engine as our NMT translation system.

For the sentence-level paraphrasing, we use a seq2seq paraphrase model which contains a bidirectional LSTM encoder and LSTM decoder together with an attention network.\footnote{https://github.com/vsuthichai/paraphraser} The model is trained on a mixed data set consisting of paraphrases from para-nmt-5m, Quora question pairs, SNLI and Semeval \cite{wieting-17-backtrans,wieting2017paranmt}. In the decoder part, we can either use a greedy search to generate a single unique paraphrase for each entire user utterance, or generate a plenty of different paraphrases via sampling from a distribution.

\subsection{Implementation Details for the four Data Augmentation Approaches}
\label{sec:supplemental}

Synonym substitution: we created four different utterances for each user utterance by randomly replacing words with their synonyms. The created data was combined with the original training data. The size of the augmented data in this way was 5 times as large as that of the original training data. 

Stop-word deletion: for this augmentation, we utilized the dictionary of stop words from NLTK toolkit and created only one copy for each user utterance and combined the additional copy with the original data. 

Translation: user utterances in original English version data were translated into Chinese, Japanese, French, German via Google Translate, and then translated back to English, thus forming four sets of data expressed in different styles. 

Paraphrasing: we generated four sets of dialogue data with the seq2seq-based paraphrase generator. 

Assembled Augmentation: we combined all data generated by the four methods above. Together, the size of the assembly augmented data is 14 times as large as that of the original data. 

The sizes of mini-batch and vocabulary for each data augmentation approach on the two datasets are shown in Table \ref{hyper parameter settings}, which are chosen according to the performance on the development set.

\begin{table*}[tb]
\caption{\label{hyper parameter settings} The sizes of mini-batch and vocabulary for the four data augmentation approaches.}
\begin{center}
\begin{tabular}{|l|c|c|l|c|c|l|}
\hline
\multirow{2}{*}{}      & \multicolumn{3}{c|}{CamRest676}                   & \multicolumn{3}{c|}{KVRET}                       \\ \cline{2-7} 
                       & Batch size      & \multicolumn{2}{c|}{Vocab size} & Batch size     & \multicolumn{2}{c|}{Vocab size} \\ \hline
Synonym Substitution   & 64              & \multicolumn{2}{c|}{800}        & 32             & \multicolumn{2}{c|}{1800}       \\ \hline
Stop-Word Deletion     & 32              & \multicolumn{2}{c|}{800}        & 32             & \multicolumn{2}{c|}{1400}       \\ \hline
Translation            & 100             & \multicolumn{2}{c|}{800}        & 32             & \multicolumn{2}{c|}{1800}       \\ \hline
Paraphrasing           & 64              & \multicolumn{2}{c|}{800}        & 64             & \multicolumn{2}{c|}{1800}       \\ \hline
Assembled Augmentation & 64              & \multicolumn{2}{c|}{800}        & 256            & \multicolumn{2}{c|}{1800}       \\ \hline
\end{tabular}
\end{center}
%\label{hyper parameter settings}
\end{table*}

\begin{table*}[tb]
\caption{\label{tab Experiment results on CamRest676 and KVRET} Experiment results on CamRest676 and KVRET.}
\begin{center}
\begin{tabular}{|l|l|c|l|c|l|}
\hline
\multicolumn{2}{|l|}{\multirow{2}{*}{}}                            & \multicolumn{2}{c|}{CamRest676}     & \multicolumn{2}{c|}{KVRET}          \\ \cline{3-6} 
\multicolumn{2}{|l|}{}                      & \multicolumn{4}{c|}{Success F$_1$}                                        \\ \hline
\multicolumn{6}{|c|}{Results from \cite{lei2018sequicity}}                                                                                            \\ \hline
\multicolumn{2}{|l|}{TSCP}                                         & \multicolumn{2}{c|}{0.834}          & \multicolumn{2}{c|}{0.774}          \\ \hline
\multicolumn{2}{|l|}{TSCP + RL}                                    & \multicolumn{2}{c|}{0.854}          & \multicolumn{2}{c|}{0.811}          \\ \hline
\multicolumn{6}{|c|}{Our implementation}                                                                                                       \\ \hline
\multicolumn{2}{|l|}{TSCP}                                         & \multicolumn{2}{c|}{0.832}          & \multicolumn{2}{c|}{0.815}          \\ \hline
\multicolumn{2}{|l|}{TSCP + RL}                                    & \multicolumn{2}{c|}{0.858}          & \multicolumn{2}{c|}{0.831}          \\ \hline
\multicolumn{6}{|c|}{Results obtained by data augmentation}                                                                                    \\ \hline
\multicolumn{2}{|l|}{Translation}                                  & \multicolumn{2}{c|}{0.869}          & \multicolumn{2}{c|}{0.842}          \\ \hline
\multicolumn{2}{|l|}{Paraphrasing}                                 & \multicolumn{2}{c|}{0.869}          & \multicolumn{2}{c|}{0.841}          \\ \hline
\multicolumn{2}{|l|}{Synonym Substitution}                         & \multicolumn{2}{c|}{0.871}          & \multicolumn{2}{c|}{0.833}          \\ \hline
\multicolumn{2}{|l|}{Stop-Word Deletion}                           & \multicolumn{2}{c|}{0.856}          & \multicolumn{2}{c|}{0.831}          \\ \hline
\multicolumn{2}{|l|}{Assembled Augmentation}                       & \multicolumn{2}{c|}{\textbf{0.879}} & \multicolumn{2}{c|}{\textbf{0.852}} \\ \hline
\multicolumn{2}{|l|}{Machine Utterance Augmentation (synonym substitution)} & \multicolumn{2}{c|}{0.775}          & \multicolumn{2}{c|}{-}              \\ \hline
\multicolumn{2}{|l|}{User + Machine Utterance Augmentation (translation)}  & \multicolumn{2}{c|}{0.822}          & \multicolumn{2}{c|}{-}              \\ \hline
\end{tabular}
\end{center}
%\label{tab Experiment results on CamRest676 and KVRET}
\end{table*}

\section{Experiments and Analyses}
%\label{Experiments and Analyses}
We conducted extensive experiments and analyses on two datasets to validate the effectiveness of the proposed methods in this section.

\subsection{Datasets and Settings}

We used two datasets: CamRest676 \cite{wen2017latent,wen2016conditional,wen2016network} and KVRET \cite{eric2017key}, both of which are manually created by crowd-sourcing workers on the Amazon Mechanical Turk platform by a Wizard-of-Oz method \cite{kelley1984iterative}. CamRest676 contains 676 dialogues in the restaurant searching domain while KVRET covers three domains: calendar scheduling, weather information and point of interest (POI) navigation.

For TSCP, the dimensionality $d$ for both hidden states and word embeddings was set to 50. Vocabulary size $V$ was 800 for CamRest676 and 1400 for KVRET. The size of mini-batch for both datasets was set 32. The model was trained with the Adam optimizer \cite{kingma2014adam}, with a learning rate of 0.003 and a decay parameter λ of 0.5. We used a learning rate of 0.0001 and decay of 0.8 for the subsequent reinforcement learning process. We used beam search strategy with a beam size of 10 on CamRest676 and greedy search strategy on KVRET. Early stopping was also performed to improve the training efficiency.

We used the Success F$_1$ score as the automatic metric for dialogue evaluation. The Success F$_1$ score estimates both the precision and recall of requested slots being successfully answered in dialogues, a metric proposed by \cite{lei2018sequicity}.

\begin{table*}[t]
\caption{\label{tab Analysis of the improvement reasons} Analysis on the Precision and Recall.}
\begin{center}
\begin{tabular}{|l|c|c|c|c|c|c|}
\hline
\multirow{2}{*}{}      & \multicolumn{6}{c|}{CamRest676}                  \\ \cline{2-7} 
                       & Success F$_1$ & Precision & Recall & TP  & FP & FN  \\ \hline
baseline               & 0.832      & 0.885     & 0.786  & 422 & 55 & 115 \\ \hline
Assembled Augmentation & 0.879      & 0.883     & {\textbf{0.875}}  & {\textbf{467}} & 62 & {\textbf{67}}  \\ \hline
\end{tabular}
\end{center}
%\label{tab Analysis of the improvement reasons}
\end{table*}

% \usepackage{graphicx}
\begin{table*}[t]
\caption{\label{sample dialogues} Dialogue examples of the baseline and data augmentation.}
\centering
\resizebox{\textwidth}{!}{%
\begin{tabular}{|l|l|}
\hline
\textbf{\textbf{Baseline}} & \textbf{\textbf{Data augmentation}} \\ \hline
\begin{tabular}[c]{@{}l@{}}\textbf{User}: Can you tell me if there is a moderately priced restaurant \\ serving spanish food anywhere in town?\\ \textbf{Machine}: la tasca serves spanish food.\\ \textbf{User}: What is the address, phone number, and area?\\ \textbf{Machine}: la tasca's phone number is 01223 464630.\\ \textbf{User}: That is it. Thank you.\\ \textbf{Machine}: I'm sorry , there are no restaurants serving spanish food.\\ \textbf{User}: Thanks again! bye.\\ \textbf{Machine}: Thank you for using the cambridge restaurant system.\end{tabular} & \begin{tabular}[c]{@{}l@{}}\textbf{User}: Can you tell me if there is a moderately priced restaurant \\ serving spanish food anywhere in town?\\ \textbf{Machine}: la tasca serves spanish food in the moderate price range.\\ Would you like their location ?\\ \textbf{User}: What is the address, phone number, and area?\\ \textbf{Machine}: Their address is 14 -16 bridge street and their phone\\ number is 01223 464630.\\ \textbf{User}: That is it. Thank you.\\ \textbf{Machine}: Have a nice day.\\ \textbf{User}: Thanks again! bye.\\ \textbf{Machine}: Have a nice day.\end{tabular} \\ \hline
\begin{tabular}[c]{@{}l@{}}\textbf{User}: I would like a restaurant in the north part of town that serves\\ australasian food\\ \textbf{Machine}: There are no restaurants in the north part of town. Would\\ you like a different type of food ?\\ \textbf{User}: How about asian oriental food?\\ \textbf{Machine}: dojo noodle bar serves asian oriental food.\\ \textbf{User}: Yes, I would like their address and phone number. \\ \textbf{Machine}: dojo noodle bar is located at 40210 millers yard city centre \\ and their phone number is 01223 363471.\\ \textbf{User}: Thank you goodbye.\\ \textbf{Machine}: I'm sorry , but dojo noodle bar is a nice day.\end{tabular} & \begin{tabular}[c]{@{}l@{}}\textbf{User}: I would like a restaurant in the north part of town that serves\\ australasian food.\\ \textbf{Machine}: I'm sorry , there are no restaurants serving australasian food.\\ Would you like me to find a different restaurant?\\ \textbf{User}: How about asian oriental food?\\ \textbf{Machine}: saigon city serves asian oriental food\\ \textbf{User}: Yes, I would like their address and phone number. \\ \textbf{Machine}: Their address is 169 high street chesterton chesterton. \\Their phone number is 01223 356555.\\ \textbf{User}: Thank you goodbye.\\ \textbf{Machine}: Have a nice day !\end{tabular} \\ \hline
\end{tabular}%
}
\end{table*}

\subsection{Results}

Table \ref{tab Experiment results on CamRest676 and KVRET} shows the experiment results on the two datasets, from which we have three findings. First, the results demonstrate that all the proposed data augmentation methods contribute to the significant improvements in F$_1$ over the basic TSCP model. Except for the stop-word deletion method, all other methods perform better than the RL-enhanced TSCP. Second, the sentence-level augmentation methods are better than the word-level methods in most cases as the former provide more variances for user utterances. Third, the assembled augmentation, which combines all data generated by the four data augmentation methods, achieve the new state-of-the-art performance on the two datasets, with more than 2 points higher than the RL-enhanced TSCP model in terms of F$_1$ score.

\subsection{Effect of Augmentation on Machine Utterances}

At each turn in a dialogue from the two  datasets, a user utterance triggers some special requests and a machine response utterance provides answers to these requests. In our previous experiments, we performed data augmentation only on user utterances.  In order to study the effect of data augmentation on machine utterances, we further carried out two experiments. One is to generate both user and machine utterances with the translation augmentation method. The other is to create copies only for machine utterances with synonym substitution. Both experiments were carried out on the CamRest676 dataset. 

Results are displayed at the bottom of Table \ref{tab Experiment results on CamRest676 and KVRET}. It is clear to observe that machine utterance augmentation seriously deteriorates the performance. The reason for this may be that data augmentation introduces both variance and noise. The variance and noise in user utterances can prevent the system from over-sensitivity \cite{niu2018adversarial}, thus making the system more robust. However, the variance and noise in machine utterances will distract the system. This resonates with the back translation that uses real target sentences and translated source sentences, widely used for seq2seq-based neural machine translation \cite{sennrich2015improving}.

\subsection{Analysis}

We took a deep look into the data to investigate how the proposed data augmentation methods improve the Success F$_1$ score that computes both the precision and recall of requested slots being correctly answered. 

The precision and recall in F$_1$ can be formulated as follows:

\begin{equation}
    F_1 = 2 \cdot \frac{precision \cdot recall}{precision + recall}
\end{equation}

\begin{equation}
    precision=\frac{TP}{TP+FP}
\end{equation}

\begin{equation}
    recall=\frac{TP}{TP+FN}
\end{equation}
where TP denotes the number of requested slots that are correctly predicted and do exist in real machine responses, FN the number of slots that exist in real responses but not answered at all, FP the number of slots being predicted but not present in real responses.

We provide the values for the precision, recall, TP, FN and FP in Table \ref{tab Analysis of the improvement reasons} for the assembled augmentation. Obviously, our method can significantly improve the recall by nearly 9 points while keeping the precision basically the same as the baseline. The reason behind the improvement of the recall is that the proposed methods substantially increases TP and decreases FN. This is because the diversity in user utterances created by data augmentation helps the dialogue system recognize more requested slots and further allows the decoder to answer these slots in  machine responses. Without data augmentation, some slots are just not detected at all in the baseline (thus a higher FN).

\subsection{Dialogue Samples}
Table \ref{sample dialogues} shows some dialogue examples generated by the model with or without data augmentation. The dialogues on the left side of the table is generated by the baseline model, while on the right side is the examples generated by the model with assembled data augmentation. Obviously, the model after our data augmentation is more robust to understand the user utterances and can produce more appropriate machine responses.

\section{Related Work}
Data augmentation has achieved great success in various tasks including computer vision \cite{krizhevsky2012imagenet}, speech recognition \cite{hannun2014deep} and text classification \cite{zhang2015character}, but is explored in a very limited way for the natural language understanding (NLU) module of traditional pipeline systems of task-oriented dialogue. \cite{kurata2016labeled} propose to augment data for the NLU module by adding noise to one single user utterance without considering its relation with other utterances. \cite{jalalvand2018automatic} introduce a technique to expand the limited in-domain data for a new spoken language understanding task. \cite{hou2018sequence} propose a data-augmentation framework to model relations between utterances of the same semantic frame in the training data. Other researchers present methods for gathering dialogue data through crowd-sourcing, e.g., via talking to myself \cite{fainberg2018talking} or MultiWOZ
\cite{budzianowski2018multiwoz}. Different from our methods, these methods either focus solely on the NLU module or rely on expensive human efforts.

\section{Conclusion and Future Work}
In this paper, we have presented four different effective methods of data augmentation for end-to-end task-oriented dialogue systems at both the word and sentence level. Empirical study on two public datasets CamRest676 and KVRET shows that data augmentation can prevent the dialogue system from the omission of key information in user utterances and significantly improve the F$_1$ score via effectively solving the problem of data scarcity. 

In the future, we intend to apply our data augmentation methods on more datasets and to explore some other efficient ways to increase the diversity of machine responses as well.


\section*{Acknowledgment}
The present research was supported by the National Natural Science Foundation of China (Grant No. 61622209 and 61861130364). We would like to thank the three anonymous reviewers for their insightful comments.



%\bibliographystyle{IEEEtran}

%\begin{thebibliography}{1}
%H.~Kopka and P.~W. Daly, \emph{A Guide to \LaTeX}, 3rd~ed.\hskip 1em %plus
%\end{thebibliography}

\bibliographystyle{IEEEtran}
\bibliography{IEEEabrv,IEEEexample}

\end{document}